\documentclass[10pt,conference]{IEEEtran}
\IEEEoverridecommandlockouts
\usepackage{cite}
\usepackage{amsmath,amssymb,amsfonts}
\usepackage{algorithmic}
\usepackage{graphicx}
\usepackage{textcomp}
\usepackage{xcolor}
\usepackage{amsfonts}
\usepackage{algorithm}
\usepackage{amsthm}
\usepackage{bm}
\usepackage{subfigure}
\usepackage{pifont}% http://ctan.org/pkg/pifont
\usepackage{booktabs, makecell}
\usepackage[utf8]{inputenc}
\usepackage[T1]{fontenc}
\usepackage{upgreek}
\usepackage{authblk}

\def\BibTeX{{\rm B\kern-.05em{\sc i\kern-.025em b}\kern-.08em
    T\kern-.1667em\lower.7ex\hbox{E}\kern-.125emX}}

\begin{document}

\newcommand{\block}[1]{
  \underbrace{\begin{matrix}1 & \cdots & 1\end{matrix}}_{#1}
}

\newcommand{\bunderbrace}[2]{%
  \begin{array}[t]{@{}c@{}}
  \underbrace{#1}\\
  #2
  \end{array}
}

\newtheorem{lemma}{\textbf{Lemma}}%[section]
\newtheorem{theorem}{\textbf{Theorem}}%[section]
\newtheorem{corollary}{\textbf{Corollary}}
\newtheorem*{remark}{\textbf{Remark}}

\newcommand{\cmark}{\ding{51}}%
\newcommand{\xmark}{\ding{55}}%

% \onecolumns

\title{Neural Reflectance Fields for Radio-Frequency \\ Ray Tracing
% {\footnotesize \textsuperscript{*}Note: Sub-titles are not captured in Xplore and
% should not be used}
% \thanks{Identify applicable funding agency here. If none, delete this.}
}

% \author{
% \IEEEauthorblockN{Haifeng Jia, Wei Lou, Hong Hu, Yichen Wei, Yifei Sun, and Yibo Pi}
% \IEEEauthorblockA{Shanghai Jiao Tong University, Shanghai, China}
% % \IEEEauthorblockN{Yibo Pi}
% % \IEEEauthorblockA{\textit{Shanghai Jiao Tong University}}
% % \and
% % \IEEEauthorblockN{Daqian Ding}
% % \IEEEauthorblockA{\textit{Beijing University of Posts and Telecommunications}}
% % \and
% % \IEEEauthorblockN{Haorui Li}
% % \IEEEauthorblockA{\textit{Dalian University of Technology}}
% % \and
% % \IEEEauthorblockN{6\textsuperscript{th} Given Name Surname}
% % \IEEEauthorblockA{\textit{dept. name of organization (of Aff.)} \\
% % \textit{name of organization (of Aff.)}\\
% % City, Country \\
% % email address or ORCID}
% }

\author[1]{Haifeng Jia}
\author[2]{Xinyi Chen}
\author[1]{Yichen Wei}
\author[1]{Yifei Sun}
\author[1]{Yibo Pi}
\affil[1]{Shanghai Jiao Tong University, Shanghai, China}
\affil[2]{East China Branch of State Grid
Corporation of China, Shanghai, China}
\affil[ ]{Email: yibo.pi@sjtu.edu.cn}

\maketitle

\begin{abstract}
Ray tracing is widely employed to model the propagation of radio-frequency (RF) signal in complex environment. The modelling performance greatly depends on how accurately the target scene can be depicted, including the scene geometry and surface material properties. The advances in computer vision and LiDAR make scene geometry estimation increasingly accurate, but there still lacks scalable and efficient approaches to estimate the material reflectivity in real-world environment. In this work, we tackle this problem by learning the material reflectivity efficiently from the path loss of the RF signal from the transmitters to receivers. Specifically, we want the learned material reflection coefficients to minimize the gap between the predicted and measured powers of the receivers. We achieve this by translating the neural reflectance field from optics to RF domain by modelling both the amplitude and phase of RF signals to account for the multipath effects. We further propose a differentiable RF ray tracing framework that optimizes the neural reflectance field to match the signal strength measurements. We simulate a complex real-world environment for experiments and our simulation results show that the neural reflectance field can successfully learn the reflection coefficients for all incident angles. 
As a result, our approach achieves better accuracy in predicting the powers of receivers with significantly less training data compared to existing approaches.

\end{abstract}

% \begin{IEEEkeywords}
% mmWave backhauling, interference graph estimation, resource allocation
% \end{IEEEkeywords}

\section{Introduction}
Ray tracing is a fundamental tool for the accurate modelling of electromagnetic (EM) wave propagation in complex environments, which enables a diverse range of applications, e.g., channel modelling~\cite{rt_channel_modelling}, antenna design~\cite{rt_antenna_design}, 5G deployment~\cite{5g_deployment}, and indoor localization~\cite{rt_indoor_localization}. Employing ray tracing in real-world environment requires a high-quality scene model including the geometry of objects and their material properties. With the advances in computer vision and LiDAR, 3D scene geometry and material estimation has become increasingly accurate and efficient~\cite{visual_slam, lidar_slam}. Material estimation in computer vision focuses on identifying the material properties, e.g.,  surface color, roughness, and metallic, that affect visual effects~\cite{Neilf}. In contrast,  ray-frequency (RF) ray tracing cares about the material reflectivity, which determines the amplitude and phase changes of the RF signal before and after reflection. Material reflection coefficient is not a constant for the given material but a function of several factors including incident angle, thickness, etc~\cite{reflection_coefficient_factors}. It is challenging to accurately measure the material reflectivity in complex real-world environments in an efficient way. 

The early work on material reflectivity measurement can be dated back to decades of years ago, where a pair of TX and RX nodes moves around the material sample to measure the attenuation of the reflected signal at different angles~\cite{simple_reflectivity_method}. The material sample size and the antenna aperture have to be carefully selected, and electromagnetic absorbers are placed behind the material sample to combat the multi-path effects. Scaling this approach to complex environments requires measuring each material in a controlled experiment separately, which is apparently not scalable. Along the same line of work, the feasibility of material identification with a single mmWave radio is studied in~\cite{material_sensing}. While this approach greatly reduces the efforts for material reflectivity measurement, it is still labor-intensive for practical use in large-scale scenes. Further, compared to millimeter wave, it is more challenging to measure material reflectivity using low frequency waves due to rich paths between TX-RX pairs~\cite{sensing_survey}.

\begin{figure*}[!t]
    \centering
        \includegraphics[scale=0.4]{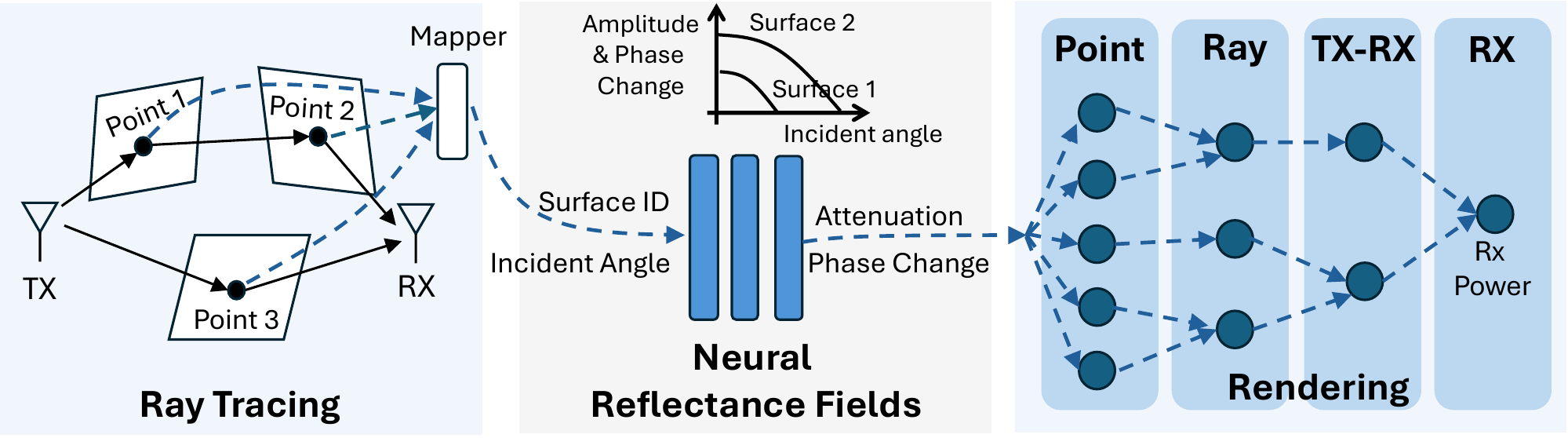}
    \caption{Our proposed RF ray tracing framework with the neural reflectance field. Our framework includes three components: (1) \textbf{Ray tracing module}, which employs current ray tracing algorithms to discover the strong rays for each TX-RX pair, maps the reflection points to their surface IDs, and steams the tuples, (incident angle, surface ID), to the neural reflectance fields. (2) \textbf{Neural reflectance field}, which takes as input the tuples and outputs the attenuation and phase change of the RF signal. (3) \textbf{Rendering module}, which composes point-level attenuation into ray-level attenuation, aggregates across multiple rays to compute the received signal for each TX-RX pair, and sums up the received signal from all transmitters to compute the receive power.}\label{fig:framework}
\end{figure*}

Scalable material reflectivity measurement is made possible with the material identification techniques in computer vision, where material types in real-world environments are first identified from the scene images and then used to infer material reflectivity. For outdoor environments, street images readily available from the Google map are used to construct a 3D material map for mmWave communications~\cite{mmSV}. Together with ray tracing, the 3D material map can be used to determine the best  direction for beamforming. However, as mentioned before, material reflectivity depends not only on material type, but also the incident angle and material thickness, which cannot be explicitly inferred from images. It is possible to compute the reflection coefficient using the Fresnel equations, but the estimated coefficients may deviate significantly from the actual ones~\cite{meas_theo_diff}.

In recent years, the neural radiance field (NeRF)~\cite{nerf} has been shown a great success in the ray tracing of light. Given a few images of the scene, it can learn a continuous representation of the scene's volumetric properties to generate new images from different angles without the need for a scene model. NeRF$^2$~\cite{nerf2} extends the neural radiance field from optics to electromagnetism by considering the reflection, diffraction, and scattering of RF signal. NeRF$^2$ represents scenes as neural radiance fields and optimize the neural networks with RF signal measurements. Given a transmitter with known position, NeRF$^2$ can accurately predict the receive power at any position when the data density is sufficient. However, NeRF$^2$ has three key drawbacks. First, its performance depends on the data density and may experience high prediction errors under low data density. Second, it fails to achieve accurate prediction in regions where sudden jumps in measurements exist even when data density is high. Most importantly, although the predicted receive power is accurate, the learned volumetric scene function cannot reflect the real radio propagation environment. These issues are inherent to data-driven approaches~\cite{rf-diffusion}.

To resolve the above issues, we propose RF ray tracing with accurate scene models. Specifically, to predict the receive power at a position, we first leverage ray tracing algorithms to discover all strong rays from the transmitter to the position, compute the receive power at each ray, and add up ray-level complex receive powers. As mentioned, an accurate scene model includes the 3D geometry of objects and their material properties. Considering the success of 3D scene reconstruction (e.g., OpenStreetMap~\cite{OpenStreetMap} and Google Map), we simply assume that the 3D scene geometry is known and focus on learning the complex material reflectivity functions, which are challenging to measure at scale. To address this challenge, we leverage the strong descriptive power of neural networks and propose neural reflectance fields for materials in real-world environments, where the neural reflectance fields are capable of learning the material reflection coefficients at different incident angles. 
For efficient and scalable measurement, we can learn the neural reflectance fields in complex outdoor environments using the receive power of user equipments (UEs) in cellular networks and the 3D scene geometry provided by OpenStreetMap. Our approach can be easily extended to indoor scenarios with 3D scene geometry constructed by popular mapping tools, e.g., SLAM~\cite{slam_survey}.

Specifically, we translate the neural reflectance fields from optics to RF domain with a complex-valued multi-layer perception (MLP), where both the amplitude and phase of RF signals are considered to model the constructive and destructive effects resulting from multipaths. The neural reflectance fields take as input the incident angle and surface ID to predict the amplitude and phase changes due to reflection of the surface. In our neural network, we compute attenuation at the ray level using the surface-level attenuation predicted by the neural reflectance fields. This allows the neural network to decompose the ray-level attenuation from the combined receive power of a UE, which is a mixture of signals from both the serving and interfering base stations (BSs). Experimental results show that our neural network excels at predicting not only the combined receive power of a UE, but also the respective attenuation for each of the communication and interference channels, which have great potentials for resource allocation and scheduling tasks.

In summary, our contributions in this paper are as follows.
\begin{itemize}
    \item We propose a RF ray tracing framework that combines traditional ray tracing with the neural reflectance field to learn the complex material reflectivity functions in real-world environments.
    \item We translate the neural reflectance field from optics to RF domain with a complex-valued MLP and optimize the neural reflectance field to match the receive power of RF signals.
    \item We demonstrate with simulations that our approach outperforms NeRF$^2$ in predicting the total receive power and that our approach can accurately predict the receive power at the ray level, making our approach valuable to a diverse range of applications.
\end{itemize}

\section{Ray Tracing with Neural Reflectance Fields}

\subsection{Framework}

We combine the general ray tracing framework with the neural reflectance field that accounts for both scene geometry and material reflectance. We decompose the scene geometry into separate surfaces and assign each surface an ID as its reference. Each surface is assumed to adopt the same material such that material reflectance can be measured at the surface level. Unlike many existing works, we do not assume that the material types of surfaces are known in advance and thus measure material reflectance separately for each surface in the scene. As shown in Fig. \ref{fig:framework}, given the scene geometry and the locations of TX and RX nodes, we can employ traditional ray tracing algorithms to discover all strong rays from TX nodes to RX nodes. For each reflection point along a ray, we obtain the incident angle of the ray and the surface ID at the reflection point. Then, the neural reflectance field regresses attenuation $\delta(\theta, s)$ from the incident angle and surface ID for all reflection points along the ray, where $\theta$ is the ray's incident angle at the reflection point and $s$ is the surface ID. The renderer takes as input the attenuations at all reflection points along a ray, the free-space path loss (FSPL), and the transmit powers of transmitters to calculate the receive powers at different locations. This framework is differentiable, allowing us to fit the neural reflectance field to match the RF signal measurements.

\subsection{Neural Reflectance Field}

\begin{figure}[t!]
\centering
    \subfigure[Attenuation vs. angle of incidence]
    {\label{fig:att_vs_aoi}
		\includegraphics[scale=0.45]{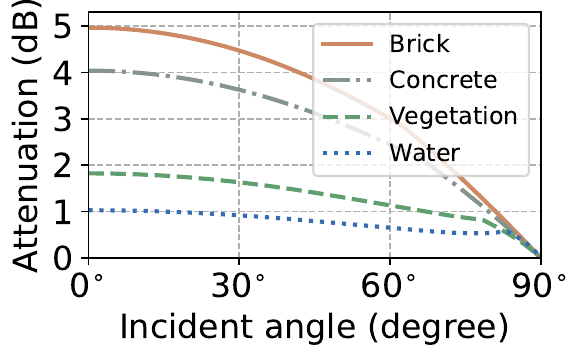}}
    \subfigure[Data density issue]
    {\label{fig:data_density_issue}
		\includegraphics[scale=0.45]{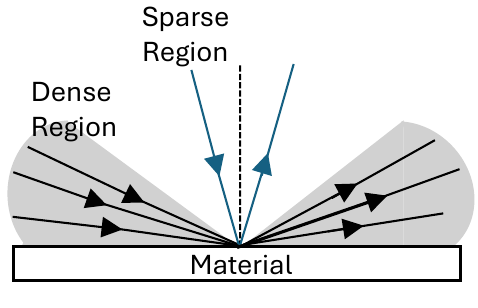}}
    \caption{Learning the relation between reflection attenuation and the angle of incidence with the neural reflectance field. Incident rays may differ in density at different angles, resulting in sparse regions to have larger learning errors for the reflection attenuation.}
    \label{fig:nonlinearity_issues_breakdown}
\end{figure}

The neural reflectance field aims to learn the amplitude and phase changes of the ray at all reflection points. Let $\Delta A(\theta, s)$ and $\Delta\Theta(\theta, s)$ denote the amplitude and phase changes of the ray incident on the $s$-th surface at an angle of $\theta$, respectively. We can represent the \emph{reflection coefficient} as $\delta(\theta, s) = \Delta A(\theta, s)e^{j\Delta \Theta(\theta, s)}$. Fig.~\ref{fig:att_vs_aoi} shows an example of the relation between reflection attenuation (amplitude change) and the angle of incidence for several typical materials in the MATLAB toolbox. The neural reflectance field can be represented as:
\begin{equation*}
    F_{\bm{w}}: (\theta, s) \mapsto \left(\Delta A(\theta, s), \Delta \Theta(\theta, s)\right)
\end{equation*}
where $\bm{w}$ indicates the neural network weights. Unlike NeRF$^2$, the neural reflectance field takes as input the incident angle(s) of the ray, because the material reflectivity depends on the incident angle. Other impacting factors to material reflectivity, e.g, thickness, are not considered in the neural reflectance field because these factors are static. To consider the multipath effects in RF domain, we need to consider both the amplitude and phase of RF signal and thus fit the reflectance field using a complex-valued neural network. We use a classic MLP architecture composed of eight fully connected layers with the ReLU activations and 256 channels, where a skip connection concatenates the input with the fifth layer's activation~\cite{nn_arch}. As a data-driven approach, the neural reflectance field experiences the data density issue as shown in Fig.~\ref{fig:data_density_issue}, where the density of incident rays varies across different angles. It is expected that the neural reflectance field can achieve better accuracy in estimating reflection attenuation in dense regions than in sparse ones. We want to ensure data density by measuring a sufficient number of RX nodes.

\subsection{RF Domain Rendering}

In the optics domain, rendering is the process to synthesize 2D images from a representation of the 3D scene. In the RF domain, we want the renderer to compute the desired features of the RF signals at given positions. In this work, we focus on the receive power of the RF signal. We consider a network with multiple transmitters, where the $i$-th transmitter transmits at the power of $P^{tx}_i$. Let $\mathcal{B}$ be the set of TX nodes and $H_{i,d}$ be the channel between TX $i$ and RX $d$. The received signal at RX $d$ is a combination of the received signal from all TX nodes, which can be written as
\begin{equation} \label{eq:received_signal}
S^{rx}_k = \sum_{i \in \mathcal{B}} \sqrt{P^{tx}_{i}} H_{i,d}. 
\end{equation}
Since the received signal is complex-valued, we compute the receive power as 
\begin{equation} \label{eq:prx}
    P^{rx}_k = | S^{rx}_k |^2,
\end{equation}
where the omnidirectional antenna is assumed with uniform gain in each direction.
The channel between the TX and the RX can be further decomposed into multiple rays. Let $\mathcal{P}_i$ be the set of reflection points along the $i$-th ray, $r_{i,k}$ be the $k$-th reflection point on ray $i$, $s(r_{i,k})$ be the surface ID of the reflection point $r_{i,k}$, and $\theta(r_{i,k})$ be the incident angle of ray $i$ at the reflection point $r_{i,k}$. We can obtain the reflection points along a ray after applying the ray tracing algorithms on the 3D scene geometry. Let $d_i$ be the length of ray $i$. According to the Friis equation, we can compute the amplitude and phase changes of ray $i$ as
\begin{equation} \label{eq:fspl}
H^{FS}_i = \frac{\lambda_c}{4\pi d_i} e^{j \frac{2 \pi d_i}{\lambda_c}},
\end{equation}
where $\lambda_c$ is the wavelength of the carrier wave and $2\pi d_i/\lambda_c$ is the phase rotation due to signal traveling from the TX to RX node. Let $\delta_k$ be the total reflection attenuation along ray $k$. We can express $\delta_k$ as
\begin{equation} \label{eq:reflection_loss}
    \delta_k = \prod_{r_{k,l} \in \mathcal{P}_k} \Delta A(\theta(r_{k,l}), s(r_{k,l})) e^{j \Delta \Theta(\theta(r_{k,l}))}.
\end{equation}
Let $\mathcal{R}_{i,d}$ be the set of rays from TX $i$ to RX $d$. Multiplying the FSPL and reflection attenuation, we can express the channel between TX $i$ and RX $d$ as
\begin{equation} \label{eq:channel}
    H_{i, d} = \sum_{k \in \mathcal{R}_{i,d}} H^{FS}_k \delta_k.
\end{equation}
Combining all the above equations, we can rewrite the receive power of RX node $d$ as in Eq.~(\ref{eq:receive_power_full}).

\begin{table*}[t]
\centering
\begin{equation}\label{eq:receive_power_full}
    P^{rx}_d 
    = \left|  
    \sum_{i \in \mathcal{B}} \sqrt{P^{tx}_{i}}
    \sum_{k \in \mathcal{R}_{i,d}} \frac{\lambda_c}{4\pi d_k} e^{j \frac{2 \pi d_k}{\lambda_c}}
    \prod_{p \in \mathcal{P}_{k}} \Delta A(\theta(r_{k,l}), s(r_{k,l})) e^{j \Delta \Theta(\theta(r_{k,l}))}
    \right|^2
\end{equation}
\hrule
\end{table*}

\subsection{Loss Function}
Let $P^{rx}_i$ be the measured receive power for RX $i$ and $\hat{P}^{rx}_i$ be the predicted RX power by the neural network for RX $i$. Suppose that we have measurements for $N$ RX nodes at different locations. We can use the following loss function to train our neural network:
\begin{equation}
    \mathcal{L} = \frac{1}{N} \sum_{i=1}^{N} | P^{rx}_i - \hat{P}^{rx}_i|^2.
\end{equation}
The weights of the neural reflectance network will be optimized to minimize the loss function. We measure the receive power in the unit of decibels, such that the weak receive power is treated equally as the strong receive power. The reason for this is that the weak receive power may consist of weak rays that are the only ones reflecting on a remote surface. We want to use as many rays as possible for measurements to avoid prediction errors due to low data density. In contrast, using milliwatts for the receive power will make the neural network inclined to optimize for the strong receive power, which is a collection of strong rays. This equates to underplay the importance of weak rays, making the surfaces solely reflected by weak rays prone to large measurement errors.

\section{Implementation Details}
\noindent\textbf{Data prerequisites.} Training the neural reflectance field requires the geometry of the 3D scene, the locations of the TX and RX nodes, and the transmit/receive powers of the TX/RX nodes at those locations. Since scenes differ in their geometries, we have to run ray tracing algorithms for each scene separately to obtain the rays for each TX-RX pair, where the number of rays can be specified as needed. We train a separate neural reflectance field for each scene. 

\noindent\textbf{One-hot encoding.} Due to the unordered and categorical characteristics of the surface IDs, we apply one-hot encoding to represent surface IDs. For a reflection point on the $s$-th surface, the input is encoded into a high-dimension vector:
\begin{equation}
    E: s \mapsto \underbrace{(0, 0, \cdots ,\overbrace{1}^{s \text{-th}}, \cdots, 0, 0)}_{\text{Number of all reflection surfaces}},
\end{equation}
where the dimension is determined by the total number of reflection surfaces.
In our experiments, the encoder is applied to the input reflection surfaces respectively. 

\noindent\textbf{Ray tracing.} Our framework includes a ray tracing module that discovers the rays between a TX-RX pair. We adopt the MATLAB ray tracing module to find rays and meanwhile obtain the coordinates of reflection points, angles of incidence at reflection points as well as the propagation distance of each ray. Shooting and bouncing rays (SBR) method \cite{sbr} works as the ray tracing algorithm, with the maximum reflection times limited to $3$ and no diffraction allowed. Thereafter, we want to map reflection points to their corresponding surfaces. We export the geometric information of all the surfaces by the Blender Python scripts, and then use ray casting algorithm to determine which surface the reflection point is on. 

\noindent\textbf{Virtual reflection points.} After running the ray tracing algorithms, we can find the rays between each pair of TX and RX nodes. The number of rays may vary across different TX-RX pairs and, similarly, the number of reflection points may vary across rays between a TX-RX pair. In order to parallelize tensor operations, we insert virtual reflection points to enforce rays to have equal number of reflection points and TX-RX pairs to have equal number of rays. We intentionally set the attenuation of virtual reflection points to zero and their incident angles to 90 degrees. Besides, the transmit power for the rays with all virtual reflection points should be set to zero. Since the attenuations of virtual reflection points are constant, their derivatives are zero in backpropagation, thereby causing no influence on the real reflection points during the training. When using the neural network for inference, we have to ensure that the virtual reflection points use the same values as in training.

\noindent\textbf{Network training configurations.} Following the practice in NeRF$^2$, we employ the Adam optimizer with an initial learning rate of $1\times 10^{-3}$ and a weight decay of $5\times 10^{-5}$. A cosine annealing scheduler is used to adjust the learning rate during the training with the maximum number of iterations $T_{max} = 10000 $ and the minimum learning rate $ \eta_{min} = 1\times 10^{-6} $. Other hyper-parameters are set as default. We train our neural network on a single NVIDIA RTX 3090 GPU. It takes $90k\sim 110k$ iterations (about 2 hours) for our network training to converge.

\section{Performance Evaluation}

\subsection{Datasets}

\begin{figure}[!t]
    \centering
        \includegraphics[scale=0.35]{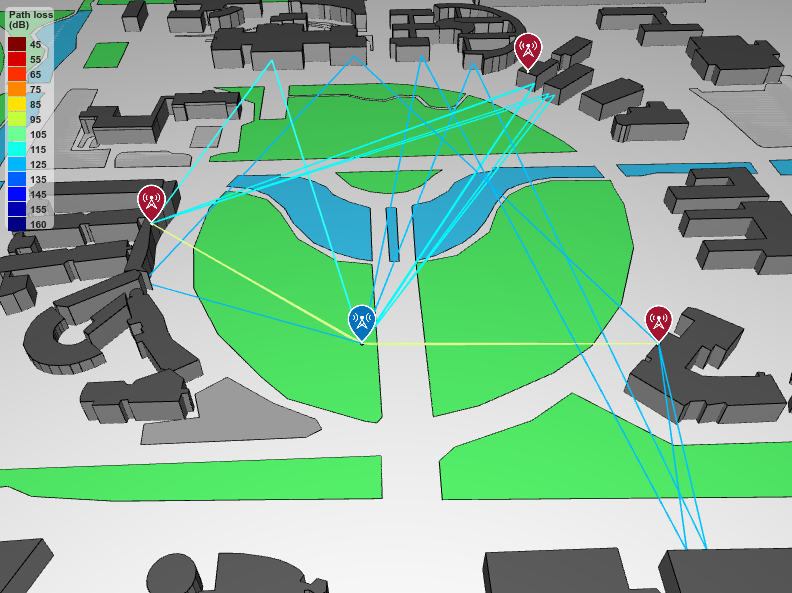}
    \caption{Campus-level 3D scene geometry from OpenStreetMap}\label{fig:dataset}
\end{figure}

We synthesize our dataset using a campus-level 3D scene obtained from the OpenStreetMap, covering an area of roughly $1\,\,km^2$ and including 923 surfaces and 4 types of materials: water, vegetation, concrete, and brick. Each material has a simulated attenuation that is linear to the angle of incidence. As shown in Fig. \ref{fig:dataset}, we deploy 3 TX nodes on the roof tops as micro BSs with a transmit power of 10 watts. Meanwhile, we randomly pick about 2000 RX nodes at the height of one meter from the ground and allow at most 21 rays for each RX and at most 3 reflections for each ray, where the number of rays is capped at 7 for each TX-RX pair. Each of the TX and RX nodes is equipped with a single omnidirectional antenna. We use 80\% of the dataset for training and the rest for testing.  

\subsection{Accuracy of Neural Reflectance Field}
\begin{figure}[t!]
\centering
    \subfigure[Accuracy vs. angle of incidence]
    {\label{fig:accu_vs_aoi}
		\includegraphics[scale=0.40]{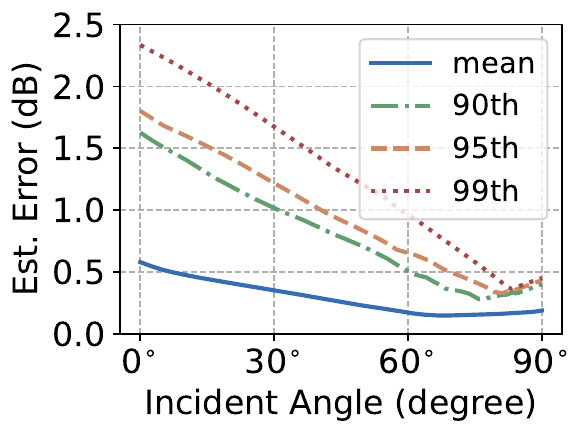}}
    \subfigure[Accuracy by material type]
    {\label{fig:accu_by_material_type}
		\includegraphics[scale=0.43]{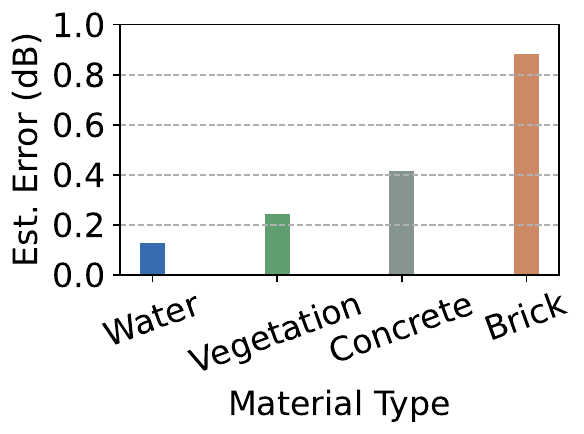}}
    \caption{Accuracy of the neural reflectance field}
    \label{fig:accu_neural_reflectance_field}
\end{figure}

\noindent\textbf{Estimation accuracy of reflection coefficients.} To evaluate the accuracy of the neural reflectance field, we want to compare its estimated reflection coefficient with the ground truth. Although our training set only includes rays with a limited number of incident angles for each surface, the neural reflectance field is capable of predicting the reflection coefficient at any angle. For each surface, we generate the estimated reflection coefficients with randomly selected incident angles and compare them with the ground-truth coefficients computed from the simulated attenuation function. We define the estimation error as $\epsilon_{\delta} = \left|10 \log_{10}(| \delta_{\text{Estimate}} | / | \delta_{\text{GroundTruth}} |) \right|$, where $\delta_{\text{Estimate}}$ is the estimated reflection coefficient and $\delta_{\text{GroundTruth}}$ is the ground truth. When there is no error, $\epsilon_\delta$ is equal to 0. Fig.~\ref{fig:accu_vs_aoi} shows the mean estimation error and the 90-th, 95-th, and 99-th percentile errors under different angles of incidence, where the angles of incidence are evenly divided into 1000 bins and the percentile is computed for errors in each bin. It can be seen that the 99-th percentile error is below 2.5 dB for all incident angles and the mean error is below 0.6 dB. We can also see that the error decreases as the angle of incidence increases and rebounds when the angle comes near 90$^{\circ}$. Based on our simulated attenuation function, reflection loss is large for small incident angles, resulting in a ray with strength weaker than others. Such weak rays have limited influence to the overall receive power and thus are likely to experience larger estimation errors. When the angle of incidence comes near 90$^{\circ}$, the reflection coefficient is very small, which tends to suffer from large estimation errors.

\noindent\textbf{Estimation accuracy by material type.} 
We further look at the estimation error by material type. Fig.~\ref{fig:accu_by_material_type} shows the 99-th percentile estimation error by material type. We can see that the estimation error increases from water to brick, where water has the smallest attenuation and brick has the largest. From water to brick, the increased attenuation weakens the ray strength, thus greatly reducing its influence to the overall receive power. As a result, the reflection coefficients of brick surfaces are the most difficult to estimate.

% \begin{figure}[!t]
%     \centering
%         \includegraphics[scale=0.5]{figs/ref-coeff-err_vs_aoi.pdf}
%     \caption{Reflection Coefficient Error with regards to Angle of Incidence}\label{fig:reflection_coefficient_error}
% \end{figure}

% \begin{figure}[!t]
%     \centering
%         \includegraphics[scale=0.5]{figs/ref-coeff-err_vs_material-type.pdf}
%     \caption{Reflection Coefficient Error of Different Material Types}\label{fig:reflection_coefficient_error}
% \end{figure}

\subsection{Comparison with Existing Approaches}

\begin{figure}[t!]
\centering
    \subfigure[Prediction error vs. data density]
    {\label{fig:pred_err_vs_density}
		\includegraphics[scale=0.43]{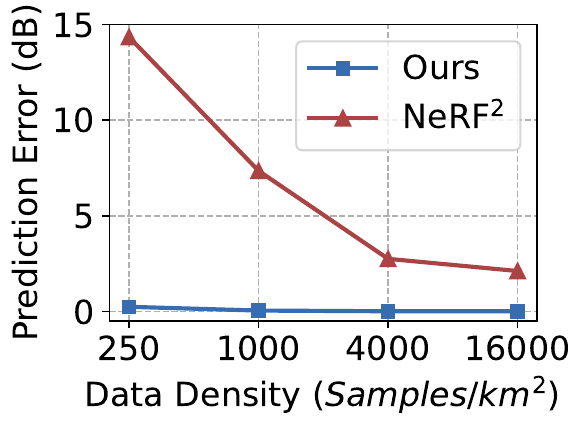}}
    \subfigure[Estimation error vs. data density]
    {\label{fig:ref_err_vs_density}
		\includegraphics[scale=0.43]{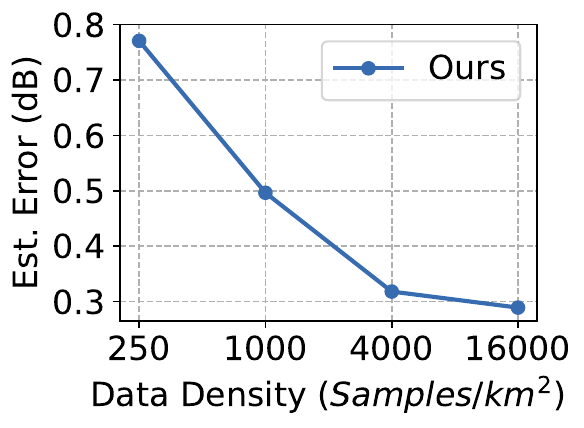}}
    \caption{Comparison with existing approaches}
    \label{fig:reflection_coefficient_error}
\end{figure}

\noindent\textbf{Prediction error.} We compare our approach with the state-of-the-art approach, NeRF$^2$, in terms of the accuracy in predicting the receive power of RX nodes. We compute the prediction error as $\epsilon_{P} = \left| 10 \log_{10}(P^{rx}_{\text{Predict}} / P^{rx}_{\text{GroundTruth}}) \right|$, where $P^{rx}_{\text{Predict}}$ and $P^{rx}_{\text{GroundTruth}}$ are the predicted and ground-truth receive powers, respectively. As shown in Fig.~\ref{fig:pred_err_vs_density}, our approach outperforms NeRF$^2$ in the mean prediction error under all data densities on the testing set. When data density is only 250 samples per km$^2$, our approach achieves a mean prediction error 14 dB less than NeRF$^2$. The prediction accuracy of NeRF$^2$ can be significantly improved by using higher data density, but its prediction error is still 2 dB higher than our approach at high data density. This is because NeRF$^2$ predicts the receive power of RX nodes at new locations with interpolation. Looking at the spatial distribution of receive powers, we find that there exist sudden jumps in receive power between nearby RX nodes, because the rays from a TX node to two close RX nodes may reflect on different reflection surfaces. This issue can be mitigated when data density is high, but is hard to entirely eliminate.

\noindent\textbf{Robustness to low data density.} 
Fig.~\ref{fig:pred_err_vs_density} shows that our approach constantly achieves a low prediction error under different data densities. This is because ray tracing allows us to group reflections on the same surface together. In other words, reflections on the same surface can be used together to train the neural reflectance field for the surface. To achieve high prediction accuracy, our approach requires sufficient measurements at the surface level, while NeRF$^2$ requires high data density for all possible locations. When the space is large as in our experiments, NeRF$^2$ needs large volume of measurements to achieve high prediction accuracy. We further evaluate our approach's capability to accurately estimate material reflection coefficients, a task that NeRF$^2$ cannot perform. We follow the same way to compute the estimation error of reflection coefficients. From Fig.~\ref{fig:ref_err_vs_density}, we can see that the neural reflectance field can achieve low estimation errors with a mean of 0.8 dB when there are only 250 samples per km$^2$ and can also benefit from higher data density. The reason for this is that more measurements at different angles improve the interpolation accuracy of the neural reflectance field.

% \begin{figure}[!t]
%     \centering
%         \includegraphics[scale=0.5]{figs/pred-err_vs_density.pdf}
%     \caption{Prediction Error}\label{fig:prediction_error}
% \end{figure}

% \begin{figure}[!t]
%     \centering
%         \includegraphics[scale=0.5]{figs/ref-coeff-err_vs_density.pdf}
%     \caption{Reflection Coefficient Error}\label{fig:reflection_coefficient_error}
% \end{figure}

% \begin{figure*}[t!]
% \centering
%     \subfigure[Received power saturation]
%     {\label{fig:saturation}
% 		\includegraphics[scale=0.40]{Images/saturation.pdf}}
%     \subfigure[Received power overshoot]
%     {\label{fig:overshoot}
% 		\includegraphics[scale=0.40]{Images/overshoot.pdf}}
%     \subfigure[RSSI sample granularity]
%     {\label{fig:RSSI_sample_accuracy}
% 		\includegraphics[scale=0.40]{Images/RSSI_sample_accuracy.pdf}}
%     \subfigure[Time sync error]
%     {\label{fig:sync_error_nonlinearity}
% 		\includegraphics[scale=0.40]{Images/sync_error_nonlinearity.pdf}}
%     \caption{Nonlinearity issues breakdown}
%     \label{fig:nonlinearity_issues_breakdown}
% \end{figure*}

\subsection{SINR and Interference Channel Prediction}

\begin{figure}[t!]
\centering
    \subfigure[SINR prediction]
    {\label{fig:CDF_SINR-err}
		\includegraphics[scale=0.42]{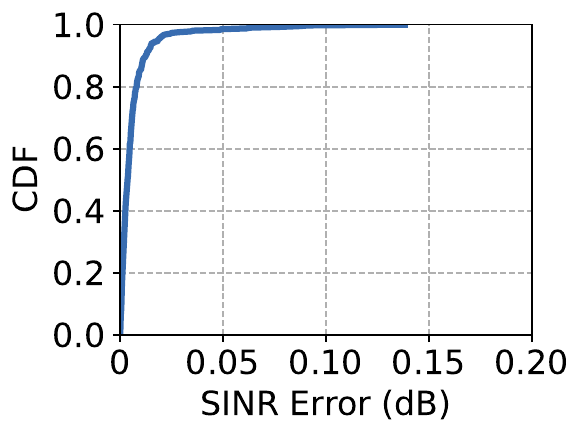}}
    \subfigure[Interference channel prediction]
    {\label{fig:CDF_interference-err}
		\includegraphics[scale=0.42]{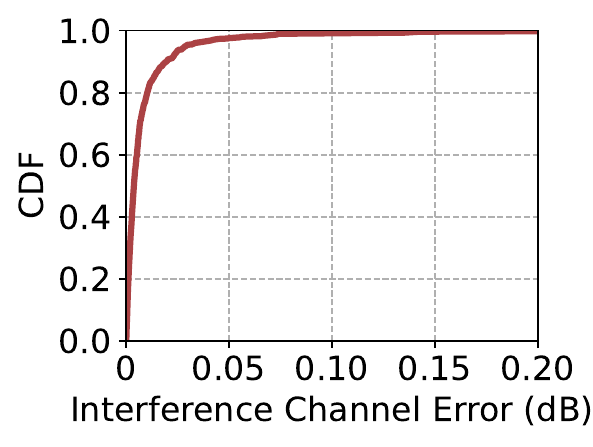}}
    \caption{Interference channel gain and SINR prediction}
\end{figure}

We evaluate the prediction error for the signal-to-interference-plus-noise ratio (SINR) and the interference channel gains. We define the \emph{interference channel error} as the difference between the predicted and ground-truth interference channel gains between TX-RX pairs, where each RX node is served by the nearest TX node and interfered by the rest. We calculate the SINR error for each RX node in the testing set and show the distribution of SINR errors in Fig.~\ref{fig:CDF_SINR-err}, where almost all RX nodes have an accurate predicted SINR deviating from the actual one less than 0.05 dB. We further decompose the superimposed interference signals at the level of TX-RX pairs. Fig.~\ref{fig:CDF_interference-err} shows that our approach can accurately predict the channel gain for each interference channel, with only less than 1\% of TX-RX pairs having a prediction error greater than 0.05 dB. Simulation results show that our ray tracing framework is powerful in predicting both communication and interference channel gains, which has great potentials to be used for resource allocation and scheduling tasks.

% \begin{figure}[!t]
%     \centering
%         \includegraphics[scale=0.5]{figs/CDF_interference-err.pdf}
%     \caption{CDF of Predicted Interference Error}\label{fig:reflection_coefficient_error}
% \end{figure}

% \begin{figure}[!t]
%     \centering
%         \includegraphics[scale=0.5]{figs/CDF_SINR-err.pdf}
%     \caption{CDF of Predicted SINR}\label{fig:reflection_coefficient_error}
% \end{figure}

\section{Conclusion}
This paper focuses on solving the problem of scalable and efficient measurement of material reflectivity for ray tracing in complex real-world environments. To address this problem, we translated the neural reflectance field that have been proved a success in optics to the RF domain to learn the reflection coefficients at any angle from the receive power of RF signals. We designed a differential RF ray tracing framework to train the neural reflectance field such that the predicted receive powers match the measured ones. In our simulations, the neural reflectance field exhibits remarkable performance in learning the material reflection coefficients at different angles for surfaces in a large scene. Furthermore, our framework demonstrated the ability to decompose the superimposed interference signals and accurately predict the gain of each interference channel, which has great potentials for resource allocation and scheduling tasks.

\section{Acknowledgement}
We appreciate the constructive feedback from the anonymous reviewers. This work was supported by the East China Branch of State Grid Corporation of China under the Grant 529924240006.

\bibliographystyle{IEEEtran}
\bibliography{IEEEabrv}

\end{document}